\documentclass{article}

\PassOptionsToPackage{numbers, compress}{natbib}
\usepackage[preprint]{neurips_2026}

\usepackage[utf8]{inputenc} 
\usepackage[T1]{fontenc}    
\usepackage{hyperref}       
\usepackage{url}            
\usepackage{booktabs}       
\usepackage{amsfonts}       
\usepackage{nicefrac}       
\usepackage{microtype}      
\usepackage{xcolor}         
\usepackage{multirow}
\usepackage{makecell}
\usepackage{amsmath}
\usepackage{wrapfig}
\usepackage[capitalize]{cleveref}
\usepackage{lipsum}
\usepackage[para]{threeparttable}

\usepackage{todonotes}
\usepackage{pifont}
\newcommand{\cmark}{\ding{51}}%
\newcommand{\xmark}{\ding{55}}%

\title{OLiVES: An Outdoor Low-Light Video Benchmark for Enhancement and Segmentation}

\author{%
  Yini Li\footnotemark[1] \\
  Visual Information Laboratory\\
  University of Bristol\\
  Bristol, United Kingdom \\
  \texttt{ub24017@bristol.ac.uk} \\
  \And
  Joanne Lin\footnotemark[1]\\
  Visual Information Laboratory\\
  University of Bristol\\
  Bristol, United Kingdom \\
  \texttt{joanne.lin@bristol.ac.uk} \\
  \AND
  David Blackham \\
  \texttt{daveblackham@mac.com} \\
  \AND
  David Bull\\
  Visual Information Laboratory\\
  University of Bristol\\
  Bristol, United Kingdom \\
  \texttt{Dave.Bull@bristol.ac.uk} \\
  \AND
  Nantheera Anantrasirichai\\
  Visual Information Laboratory\\
  University of Bristol\\
  Bristol, United Kingdom \\
  \texttt{n.anantrasirichai@bristol.ac.uk} \\
}

\begin{document}

\footnotetext[1]{These authors contributed equally to this work.}

\maketitle

\begin{abstract}
  Low-light video processing remains an underexplored problem in computer vision, primarily due to the scarcity of high-quality ground-truth data. Existing public datasets exhibit several limitations, including restricted motion diversity, limited scene coverage, weak degradations, reliance on auxiliary modalities such as event data, and misalignment between low-light inputs and ground truth. Furthermore, current datasets typically address either low-light enhancement or scene understanding, whereas these tasks should be considered jointly. We introduce \textbf{OLiVES}, a benchmark dataset for outdoor low-light videos with dynamic scenes, designed to support both low-light video enhancement (LLVE) and video object segmentation (VOS). The dataset addresses key content gaps by incorporating a diverse range of natural textures, animals, occlusions, dynamic activities, and close-up scenes. To the best of our knowledge, OLiVES is the first outdoor low-light video dataset captured using professional cinema-grade cameras. It provides over 25,000 aligned normal- and low-light frame pairs, along with over 46,000 VOS annotations. We benchmark state-of-the-art LLVE and VOS methods on OLiVES and show that training on it improves performance on both tasks. The OLiVES project page is available at: \url{https://anonymous-OLiVES.github.io/}
\end{abstract}

\section{Introduction}
\label{sec:introduction}
Enhancing the visibility and understanding of scenes captured under low-light conditions is highly desired due to its applicability to autonomous vehicles, surveillance, wildlife filmmaking and conservation, etc. While many computer vision methods show strong performance in well-lit scenes, their performance degrades significantly in low-light conditions~\cite{chen2019seeing, wang2021sdsd, lin2024BVIRLV, lin2026elvis}.
There are many reasons for this reduction in performance including noise, blur, low brightness/contrast and computational resources required to process videos.

Low-light video enhancement (LLVE) has emerged as a key task to address these challenges, aiming to improve perceptual quality or serve as a pre-processing step for downstream tasks. Low-light video enhancement provides an added challenge when compared to its image counterpart, low-light image enhancement (LLIE), due to the importance of temporal consistency. One major challenge in constructing LLVE dataset is the acquisition of paired low-light frames and corresponding ground-truth (GT). Most existing datasets adopt physical alignment methods to overcome this challenge, e.g. dual-camera systems~\cite{chen2026evlight++} or repeated shooting of the same scene~\cite{lin2024BVIRLV}. However, due to the constraints of the capture setup, the recorded videos are typically limited to either static scenes or scenes containing no moving objects. As a result, these datasets fail to adequately represent the complex motion patterns commonly encountered in real-world low-light environments. Furthermore, we observe that several existing datasets exhibit low degradation levels, e.g. DID~\cite{Fu2023DID} which simply reduces ISO to capture low-light videos.

Video object segmentation (VOS) is a further challenging downstream task whose objective is to segment and track target objects across frames. While numerous VOS benchmarks have been released~\cite{Perazzi2016davis16, PontTuset2017davis17, xu2018youtubevos18, yang2019youtubevos19, Ding2023MOSE, Ding2025MOSEv2}, the vast majority focus on well-lit conditions; to our knowledge, only~\citet{Li2024LLEVOS} has explored VOS in low-light through their LLE-VOS dataset. One of the main limitations of this dataset is that it is designed to leverage event data, which is not captured by most cameras, thus restricting its real-world applicability.
More broadly, downstream tasks in low-light conditions remain under-studied, in large part because of the scarcity of training data. A common workaround is to apply a low-light enhancement step before feeding frames into models trained on predominantly well-lit datasets. However, despite recent progress in enhancement techniques~\cite{guo2020zerodce, Cai2023retinexformer, Jiang2024ECCVLightenDiff, li2026TempRetinex}, this two-stage pipeline often hinders downstream performance. Enhancement methods improve perceptual quality but degrade the semantic cues that downstream models rely on~\cite{chen2023lis, Li2024LLEVOS, lin2025wnlb, lin2026elvis}. Moreover, existing studies indicate that jointly optimizing low-level enhancement and high-level semantics is beneficial to both tasks, improving restoration quality while simultaneously enhancing downstream performance~\cite{2022_1cac8326, Zheng2022SGZSL, liu2025jdatt, Gu2025lowlightfoco, Hill2026DMAT, lin2026elvis}.
To the best of our knowledge, the only low-light video benchmark for both enhancement and a downstream task (semantic segmentation) is SDE~\cite{chen2026evlight++}. Howevever, SDE has notable limitations: object motion is restricted (due to its setup) and the annotations are generated using off-the-shelf deep learning models without manual verification.


\begin{figure}
    \centering
    \includegraphics[width=1\linewidth]{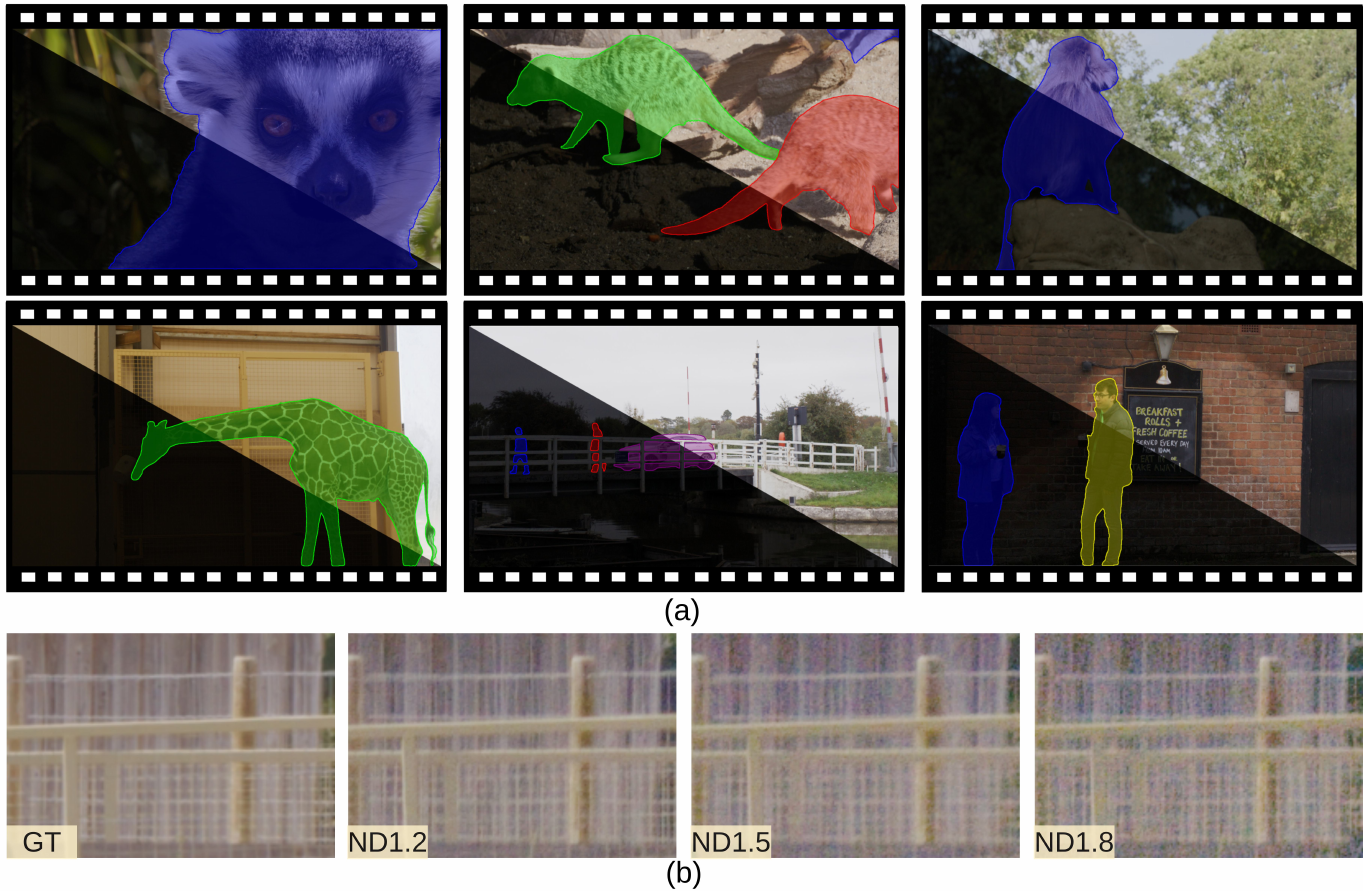}
    \caption{Examples of the OLiVES dataset. (a) Low-light videos with their corresponding normal-light ground truth and object segmentation masks. Each scene is provided at multiple light levels, (b) illustrating the noise levels exhibited in the low-light videos when using different ND filters (histogram-matched to GT for comparison).}
    \label{fig:dataset_example}
    \vspace{-20pt}
\end{figure}

To address these challenges, we propose a new low-light benchmark, \textbf{O}utdoor \textbf{L}ow-L\textbf{i}ght \textbf{V}ideos for \textbf{E}nhancement and \textbf{S}egmentation (\textbf{OLiVES}), supporting joint LLVE and VOS tasks on real-world low-light videos. Example videos are shown in \cref{fig:dataset_example}.
OLiVES provides 25,254 registered normal/low-light frames for low-light video enhancement and 46,366 pixel-wise segmentation annotations for the low-light VOS task. Unlike existing datasets that treat these tasks independently, our benchmark enables a unified evaluation of restoration and downstream understanding, encouraging research for improving both perceptual and semantic quality.
To capture diverse lighting conditions, including extreme low-light scenarios, we employ three Neutral Density (ND) filters of varying transmittance while keeping ISO and exposure time fixed. Crucially, OLiVES is captured with professional cinema-grade cameras, yielding high-fidelity footage under genuine outdoor lighting. The resulting scenes feature animals, natural textures, foliage, and dynamic outdoor activity. This opens up benchmarking and model development for application areas underserved by existing datasets, such as natural history filmmaking, wildlife monitoring, and nighttime surveillance.

\section{Related work}
\label{sec:background}
\begin{table}
    \centering
    \caption{Comparison of low-light video enhancement datasets.}
    \setlength{\tabcolsep}{4.5pt}

    \resizebox{1.0\linewidth}{!}{
    \begin{tabular}{lccccccc}
    \toprule
    \textbf{Dataset} & \textbf{Modality} & \textbf{Frames} & \makecell{\textbf{Cam.}\\\textbf{Motion}} & \makecell{\textbf{Obj.}\\\textbf{Motion}} & \textbf{Resolution} & \makecell{\textbf{VOS}\\\textbf{ann.}} & \textbf{Device} \\
    \midrule
    DRV~\cite{chen2019seeing}       & Raw        & 22,220 & Static & Static & $5496\times3672$  & \xmark & Sony RX100 VI\\
    SDSD~\cite{wang2021sdsd}        & RGB        & 37,500 & Dynamic & Static & $1920\times1080$  & \xmark & Canon 6D Mark II\\
    LLRVD~\cite{Fu2022LLRVD}        & Raw        & 4,086  & Dynamic & Dynamic & $2600\times1400$  & \xmark & Sony $\alpha$7R IV\\
    BVI-RLV~\cite{lin2024BVIRLV}    & RGB        & 31,800 & Dynamic & Static & $1920\times1080$  & \xmark & Sony $\alpha$7S II\\
    SDE~\cite{chen2026evlight++}    & RGB/Event  & 31,477 & Dynamic & Static & $346\times260$    & \xmark & DAVIS 346\\
    
    DID~\cite{Fu2023DID}            & RGB        & 41,038 & Dynamic & Static & $2560\times1440$  & \xmark & \makecell{5 Camera Models}\\
    \midrule
    \textbf{OLiVES (Ours)}    & RGB        & 25,254 & Dynamic & Dynamic & $1920\times1080$ & \cmark & RED V-RAPTOR  \\
    \bottomrule
    \end{tabular}}
    \label{tab:comparison_stats_llve}
\end{table}

\begin{table}
    \centering
    \caption{Comparison of video object segmentation (VOS) datasets.}
    \resizebox{1.0\linewidth}{!}{
    \begin{tabular}{lccccccc}
    \toprule
    \textbf{Dataset} & \textbf{Light} & \textbf{Frames} & \textbf{Categories} & \textbf{Objects} & \textbf{Annotations} & \textbf{New objects} \\
    \midrule
    DAVIS 2016~\cite{Perazzi2016davis16} & Normal & 3,440 & - & 50 & 3,455 & \xmark \\
    DAVIS 2017~\cite{PontTuset2017davis17} & Normal & 6,208 & - & 205 & 13,543 & \xmark  \\
    YouTube-VOS~\cite{xu2018youtubevos18} & Normal & 120,532 & 94 & 7755 & 197,272 & \cmark  \\
    MOSEv1~\cite{Ding2023MOSE} & Normal & 130,149 & 36 & 5,200 & 431,725 & \xmark  \\
    MOSEv2~\cite{Ding2025MOSEv2} & Mixed & 468,251 & 200 & 10,074 & 701,976 & \xmark  \\
    LLE-VOS~\cite{Li2024LLEVOS} & Low & 5,600 & - & 120 & 9,606 & \xmark  \\
    \midrule
    \textbf{OLiVES (Ours)} & Low & 22,893 & 10 & 715 & 46,366 & \cmark  \\
    \bottomrule
    \end{tabular}}
    \label{tab:comparison_stats_vos}
\end{table}


\paragraph{Low-light video enhancement.}
Low-light video enhancement is a restoration task that aims to improve perceptual fidelity. It involves several steps such as removing noise and blur degradations, increasing brightness and contrast, and enhancing colors. Several works~\cite{wang2021sdsd, Fu2023DID, lin2024BVIRLV, chen2026evlight++} have proposed benchmarks which provide low-light video frame inputs, and aligned normal-light video frames for ground truth. \cref{tab:comparison_stats_llve} summarizes existing datasets and their characteristics. However, each has notable limitations:

DRV~\cite{chen2019seeing} (also known as SMID) is a RAW low-light video dataset with aligned short/long-exposure pairs from a single device with ground truth restricted to static scenes.
SDSD~\cite{wang2021sdsd} is a public low-light video dataset captured with a mechatronic system, but suffers from spatial misalignment between normal/low-light pairs and inconsistent scenarios.
DID~\cite{Fu2023DID} is the largest LLVE dataset by frame count, using five cameras to capture diverse degradations. However, it synthesizes low-light by reducing ISO under well-lit conditions, limiting realism and noise modelling, and its gimbal setup restricts motion to linear trajectories.
LLRVD~\cite{Fu2022LLRVD} is a low-light RAW dataset with richer motion, obtained by recording videos displayed on a Dell U2720QM monitor; however, the lighting is not physically realistic, and the dataset contains only 4,086 pairs.
BVI-RLV~\cite{lin2024BVIRLV} introduces non-uniform motion via a programmable dolly and mitigates misalignment through repeated capture and frame selection, but is limited to controlled indoor scenes with static objects, lacking outdoor scenarios and real motion.
SDE~\cite{chen2026evlight++} uses a robotic arm to capture more diverse motion than prior methods and records scenes with a DAVIS 346 event camera before applying an ND8 filter for low-light sequences. However, its low resolution (346×260) limits effectiveness for detail restoration.

These benchmarks all share a common limitation: their data capture process requires re-recording scenes under different lighting conditions, thus restricting the scene motion, due to the need for accurate frame alignment. The consequence of this is that these datasets do not exhibit heavy motion blur from moving objects; a key degradation found in low-light videos which compromises perceptual fidelity and performance in computer vision tasks.
\cref{tab:comparison_stats_llve} shows that OLiVES is the only one to combine dynamic camera and object motion with VOS annotations.

\paragraph{Video object segmentation.}
Video object segmentation (VOS) is a high-level computer vision downstream task which extends the object tracking task by producing segmentation masks of objects of interest rather than bounding boxes. This provides a fine-grained understanding of the shape of the objects.
Semi-supervised VOS is typical, requiring additional data to guide the method to correctly segment the object; this is most often a mask of the first frame where an object appears.
Popular VOS benchmarks such as DAVIS~\cite{Perazzi2016davis16, PontTuset2017davis17}, YouTube-VOS~\cite{xu2018youtubevos18, yang2019youtubevos19}, MOSEv1~\cite{Ding2023MOSE} all provide annotations for the mask-initialization semi-supervised setting. Most of these benchmarks only consider objects located in the first frame except for YouTube-VOS which adds an additional difficulty to the task by having videos where new objects appear in later frames.

While these datasets are valuable for real-world applications, they provide normal-light videos. Only MOSEv2~\cite{Ding2025MOSEv2} includes low-light scenes, but does not explicitly separate them, grouping them with other challenging conditions such as underwater scenes and adverse weather.
\citet{Li2024LLEVOS} introduce LLE-VOS, a low-light VOS benchmark with aligned normal/low-light RGB frames, event data, and annotations. While it addresses a gap in the low-light domain, it has several limitations. First, performance relies heavily on event data, which is unavailable in standard cameras, limiting generalisability to RGB-only settings. Second, the low resolution (DAVIS 346) restricts fine-grained segmentation. In addition, the frames exhibit an orange colour cast from the capture process, potentially biasing model behaviour.
The statistics of these VOS datasets are summarized in~\cref{tab:comparison_stats_vos}.  Only our dataset and YouTube-VOS support objects that emerge after the first frame.

\section{The OLiVES Dataset}
\label{sec:dataset}

\begin{figure}
    \centering
    \includegraphics[width=\linewidth]{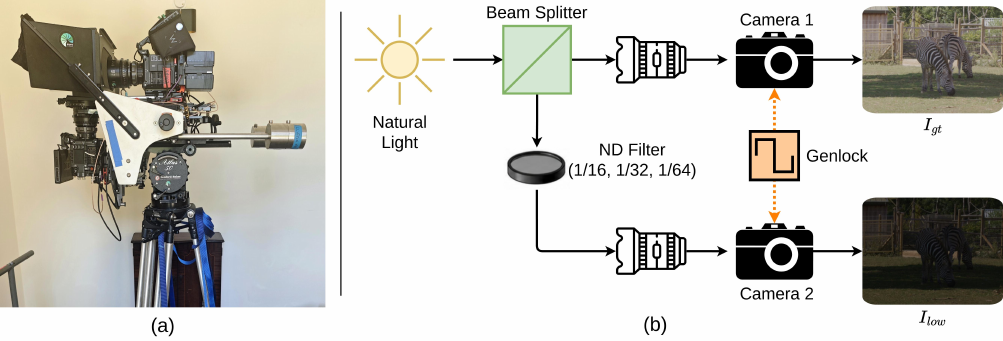}
    \caption{Capture setup. (a) Our beam-splitter rig with two genlocked cinema cameras mounted on a shared tripod. (b) Schematic of the optical path: incoming light is split between Camera 1, which records the normal-light ground truth $I_{gt}$, and Camera 2, which records the low-light frame $I_{low}$ through one of three ND filters (1/16, 1/32, 1/64). Genlock ensures temporal synchronization between the two streams.}
    \label{fig:device}
\end{figure}

\subsection{Dataset Collection}

\paragraph{Dual-exposure camera system.}
~\cref{fig:device} shows our capture setup, where a Genlock timecode system enables temporally synchronized low-light and ground-truth videos in a dual-exposure camera system. Two cameras are mounted on a Harrier beam-splitter rig and a GFM camera dolly and jib to enable smooth, controlled motion without affecting camera alignment. The beam-splitter rig ensures precise physical alignment and divides incoming light equally into two optical paths via a high-quality 50/50 half-silvered mirror, allowing both cameras to capture identical images.
Neutral density (ND) filters (Formatt Hitech Firecrest 1.2, 1.5, and 1.8), corresponding to transmittance ratios of 1/16, 1/32, and 1/64, are inserted into one path to simulate low-light conditions. These filters provide near-perfect light attenuation without noticeable color shift. The other path records the corresponding normal-light reference. A monitor connected to the cameras is used for real-time footage inspection.
For high-fidelity image acquisition, we use two professional RED V-RAPTOR cameras paired with Zeiss Prime lenses. Following standard cinematography practice, we fix the ISO at 800, the frame rate at 25 fps, the aperture at F8 to maintain sufficient depth of field in outdoor scenes, and the shutter speed at 1/100 s to reduce motion blur. All other parameters are kept at their default settings. This setup represents a practical upper bound achievable with current imaging technology.

\paragraph{Video scenes.}
Existing low-light video datasets~\cite{wang2021sdsd, Fu2023DID, lin2024BVIRLV, chen2026evlight++} predominantly contain scenes with limited motion and close-up fine textures, due to constraints in their capture setups. To address this limitation, we deliberately capture scenes with richer and more diverse dynamics. In particular, to record natural wildlife behavior, we film animals, including lemur, gelada, meerkat, giraffe, and zebra in open zoo (Bristol Zoo Project by Bristol Zoological Society, UK) enclosures. We further include outdoor scenes featuring building textures and dynamic elements such as flags, pedestrians, and vehicles, with complex scenarios including occlusions and object interactions. 
With the support of professional photographers, our dataset spans a wide range of dynamic outdoor scenarios, including humans, animals, and less commonly studied objects such as flags and balls. The captured high-resolution RAW data are first converted to 4096$\times$2160 RGB images using the official REDCINE-X PRO platform with default image signal processing settings. We then crop 1920$\times$1080 regions of interest and divide long recordings into short clips of 3 to 10 seconds.

\subsection{Dataset Registration}
\begin{figure}
    \centering
    \includegraphics[width=\linewidth]{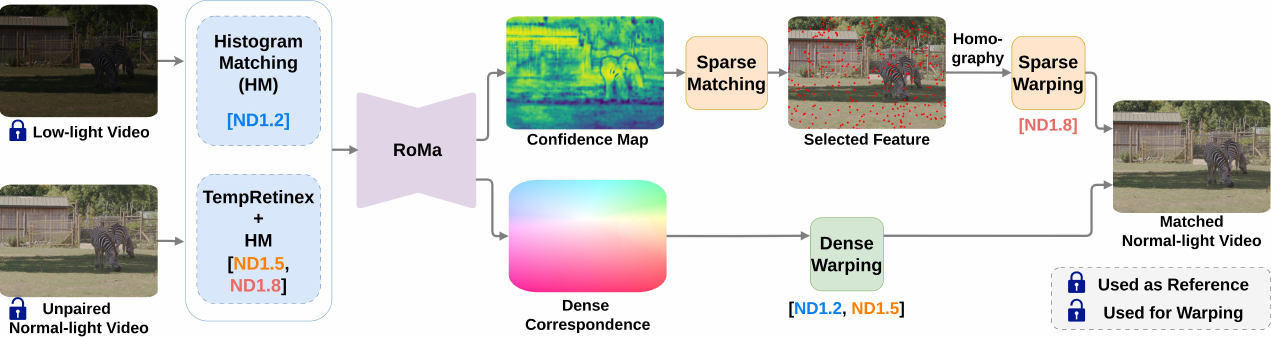}
    \caption{Registration pipeline of the OLiVES dataset.}
    \label{fig:flowchart_dataset_collection}
    \vspace{-10pt}
\end{figure}

Although Genlock ensures temporal synchronization between the two cameras, small spatial misalignments may remain between paired frames due to imperfections in beam-splitter geometry and lens calibration. To address this, we estimate dense pixel-wise correspondences using finetuned RoMa~\cite{edstedt2024roma}, followed by a warping procedure to align each normal-light frame to its corresponding low-light frame, as illustrated in~\cref{fig:flowchart_dataset_collection}. The low-light frames are kept unchanged, preserving their original distortion characteristics. The resulting transformations are limited to minor geometric corrections and do not modify photometric or structural content, thereby maintaining the validity of the normal-light frames as reference data.

\begin{figure}
    \centering
    \includegraphics[width=\linewidth]{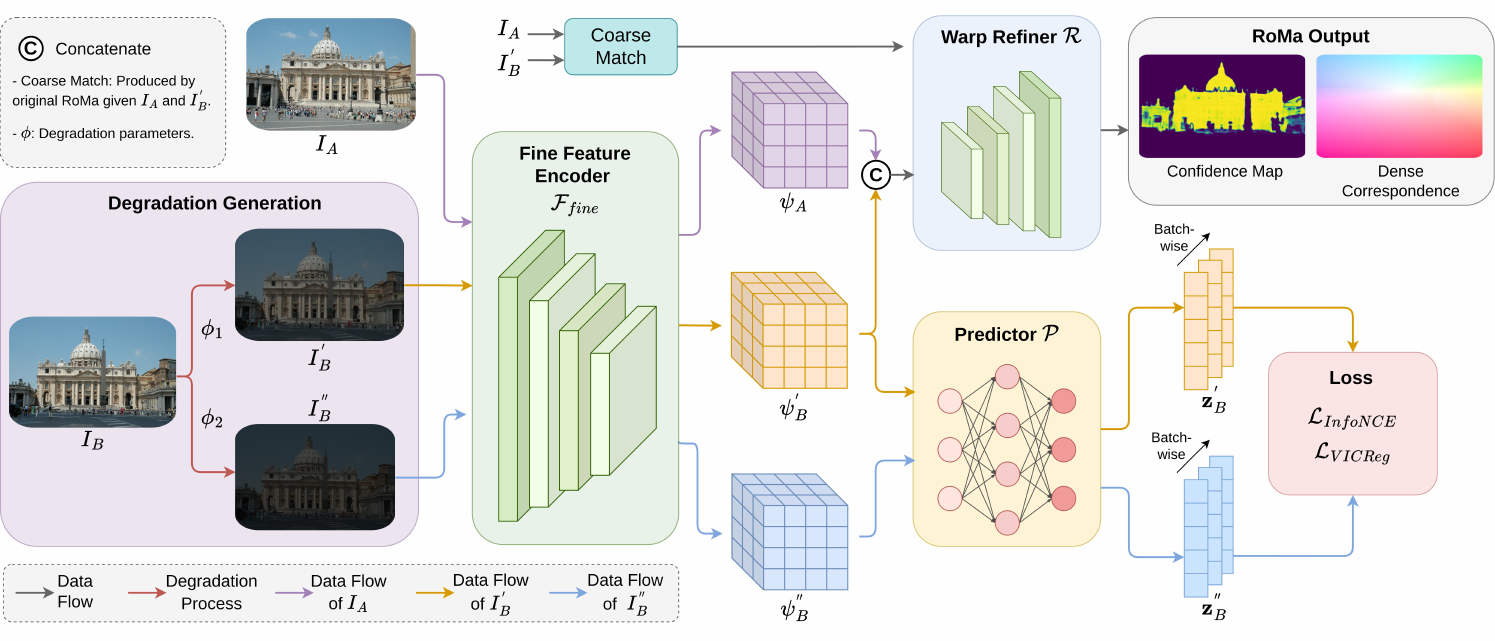}
    \caption{Finetuning the fine feature encoder of RoMa with synthetically degraded low-light pairs and a contrastive objective to improve robustness to low-light degradations.}
    \label{fig:roma}
    \vspace{-10pt}
\end{figure}

\paragraph{RoMa finetuning.}
RoMa~\cite{edstedt2024roma} produces high-quality dense correspondences along with confidence maps under standard conditions, but directly applying the pre-trained model to our dataset leads to noticeable artifacts. This mainly arises from (i) severe domain shift between normal and low-light images, and (ii) noise and detail loss in low-light input.
To improve robustness, we finetune RoMa using synthetically degraded data that mimics real low-light conditions as shown in~\cref{fig:roma}. We leverage the synthetic low-light pipeline proposed by~\citet{lin2026elvis} to create synthetic low-light pairs with degradations from our captured low-light frames, and use these synthetic frames for finetuning RoMA. In addition to this, we use contrastive learning to enforce degradation invariant feature representations. We provide further details of the finetuning process in~\cref{sec:suppmat_roma}.

\paragraph{Image Alignment.}
As low-light videos from different ND filters exhibit distinct statistical characteristics, we tailor registration strategies as in~\cref{fig:flowchart_dataset_collection}. 
For ND1.2, where details are largely preserved, we use histogram matching (HM) to reduce the brightness gap, followed by RoMa-based dense warping for alignment. We bypass enhancement for ND1.2 to keep the detail matching as accurate as possible. 
For ND1.5, which exhibits stronger noise, we first apply TempRetinex~\cite{li2026TempRetinex} for enhancement and denoising, followed by HM and dense warping. 
For ND1.8, where severe noise and color distortion compromise dense matching, we instead select reliable matched feature points using the confidence maps, and estimate a homography by RANSAC for global alignment, effectively avoiding local artifacts while maintaining alignment stability.

\subsection{VOS Benchmark Setup}
\paragraph{Video clip preparation.}
In order to provide similar lengths of videos to existing VOS benchmarks, we first randomly split the videos into smaller video clips between 40 and 100 frames. Next, we filtered the clips that were unsuitable for the VOS task (e.g. clips with no objects to track).
Unlike LLE-VOS~\cite{Li2024LLEVOS}, we include clips with emerging/disappearing objects in our benchmark for additional difficulty. This also allows for a more practical applicability as objects cannot be guaranteed to remain within the frame.
For the train/validation/test splits, we used the ratios 40\%/30\%/30\%. We ensured that the test split remained the same as the test split for the LLVE task, while for the train/validation splits, we randomly selected clips from the LLVE train split for the respective VOS splits.

\paragraph{Annotations.}
We provide annotations for 10 categories: \texttt{person}, \texttt{giraffe}, \texttt{zebra}, \texttt{monkey}, \texttt{meerkat}, \texttt{lemur}, \texttt{boat}, \texttt{car}, \texttt{flag}, \texttt{ball}.
For the training set, we first created pseudo ground truth masks by prompting SAM2~\cite{ravi2025sam2} using bounding boxes of the objects of interest onto the aligned normal-light videos. However, we found that the annotations generated with SAM2 are imperfect and thus we decided to refine a subset of the train set manually. As refining over 35K annotations is extremely time-consuming, especially for our videos where we capture difficult textures like animal fur, we randomly select 33\% of the train videos for refinement; balancing the trade-off between manual-labor and effectiveness. The refinement process was case-by-case and involved using several procedures including: applying SAM2 again on later frames to reduce the mask drift in later frames, prompting SAM2 with a mask manually drawn for better guidance, and manually editing the generated masks.
We implemented the same refinement approach for the validation set to reduce the inaccuracies from SAM2.
For the test set, we hired annotators to manually annotate the objects in the videos, referring to the matched normal-light frame for better visibility. To guide the annotators, we provided the first segmentation mask of every object in the test set. During this manual annotation process, we employed an iterative feedback loop by regularly reviewing and refining the annotations, to ensure high-quality masks. See~\cref{sec:suppmat_vos_annotators} for details on the annotation process.

\subsection{Dataset Statistics}

\begin{figure}
    \centering
    \includegraphics[width=0.95\linewidth]{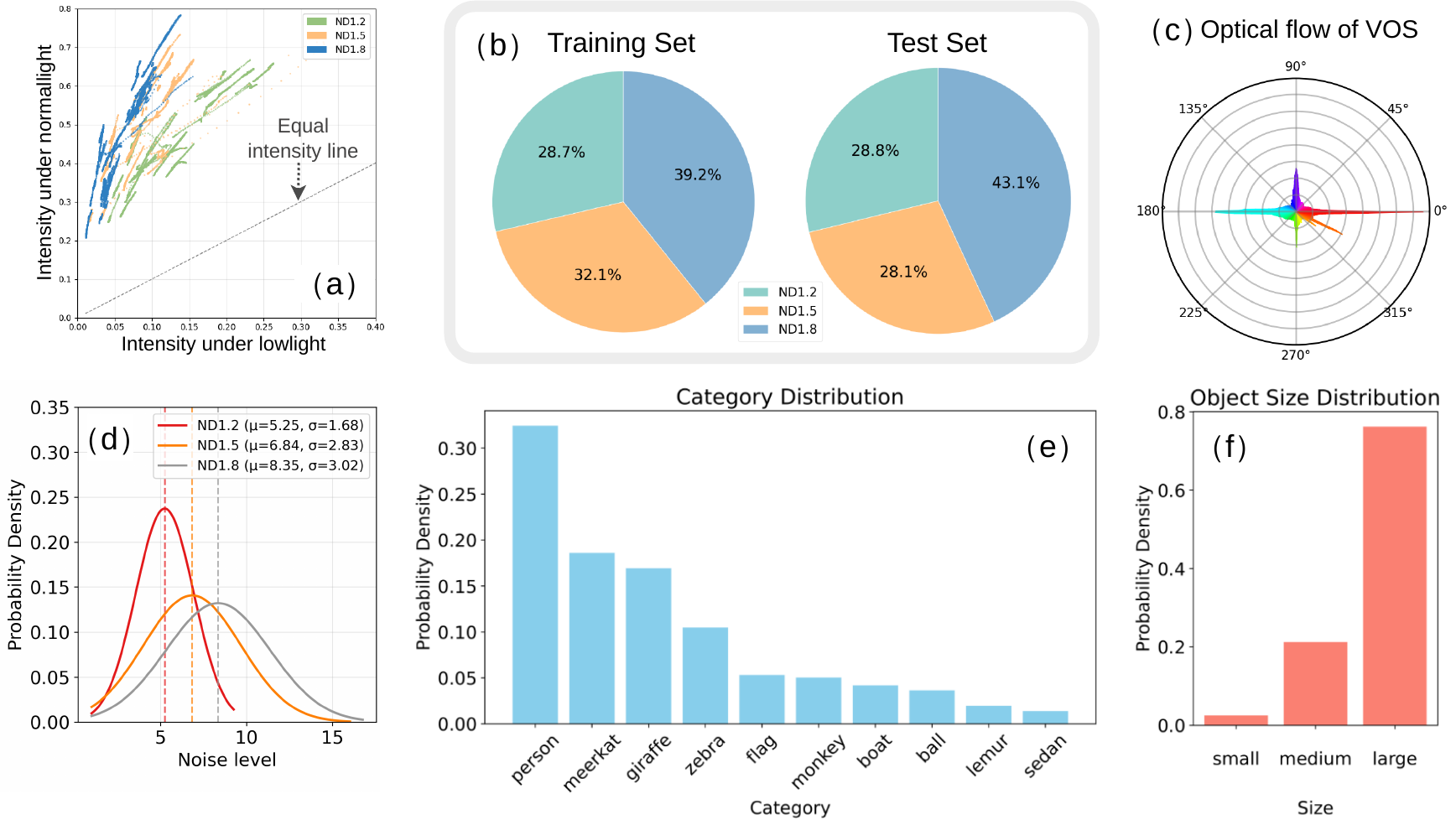}
    \caption{Dataset statistics overview. (a) Normalized pixel intensities for normal/low-light pairs, (b) ND filter ratios, (c) optical flow of VOS, (d) noise level distributions, (e) object category and size distributions.}
    \label{fig:dataset_stats}
    \vspace{-10pt}
\end{figure}

\paragraph{Composition.} OLiVES comprises 137 video sequences (3–10 seconds each), totalling 25,254 frames. We split the dataset by scene into a training set (96 sequences, 17,890 frames) and a test set (41 sequences, 7,364 frames) to avoid overlap while maintaining similar content distributions across splits. 

\paragraph{Intensity distribution.} \Cref{fig:dataset_stats}(a) illustrates the normalized pixel intensity statistics for low-light frames under three ND filters (ND1.2, ND1.5, ND1.8) and their normal-light counterparts. The low-light intensity mainly covers a range of 0.01-0.27, while the  normal-light one demonstrates a broad range of brightness ranging from 0.2 to 0.8. As shown in~\cref{fig:dataset_stats}(b), the brightness distributions are also consistent between the training and testing sets. Specifically, the ratios of ND1.2, ND1.5, and ND1.8 are 28.7\%:32.1\%:39.2\% in the training set and 28.8\%:28.1\%:43.1\% in the test set.

\paragraph{Noise characteristics.} To quantify the noise level, we compute the absolute difference $mean(||I_{hm}-I_{gt}||_{1})$ of each paired frames, where the $I_{hm}$ is the HM output of $I_{low}$. We fit the noise levels by normal distribution for better visualization in~\cref{fig:dataset_stats}(d). As the ND stop increased, the noisy degree increased (noise level mean from 5.25 to 8.35, standard deviation from 1.68 to 3.02), indicating more challenging conditions in extreme darkness of OLiVES.

\paragraph{Motion characteristics.}
To analyze the motion characteristics of the dataset, we calculate the optical flow for all video frame pairs and statistically analyze the motion direction distribution of the objects. As shown in~\cref{fig:dataset_stats}(c), objects exhibit motion in all directions, with horizontal motion being dominant, while also covering a rich variety of non-dominant directional motions.
Regarding motion amplitude, the average inter-frame displacement of the objects is 8.3 pixels, with a variance of 139.6; while the average displacement of entire image is 7.7 pixels, with a variance of 275.7. The variance indicates that the dataset contains both small and large motions, reflecting diversity in motion scale.

\begin{wraptable}{r}{7cm}
    \vspace{-10pt}
    \caption{Summary of OLiVES VOS benchmark for the train, validation and test set.}
    \centering
    \resizebox{1.0\linewidth}{!}{
    \begin{tabular}{lcccc}
    \toprule
    \textbf{Split} & \textbf{Videos} & \textbf{Frames} & \textbf{Objects} & \textbf{Anns.} \\
    \midrule
    Train & 129 & 8,826 & 275 & 17,901 \\
    Validation & 99 & 7,012 & 196 & 12,820 \\
    Test & 102 & 7,055 & 244 & 15,646 \\
    \bottomrule
    \end{tabular}}
    \label{tab:summary_vos}
\end{wraptable}

\paragraph{VOS annotation statistics.}
\Cref{tab:summary_vos} summarizes the number of objects and annotations in our dataset. We provide statistics for number of videos, frames, objects and annotations. The distributions of object classes and sizes are shown in~\cref{fig:dataset_stats}(e) and (f). We categorize object sizes based on the pixel area of their segmentation masks, following the thresholds defined by~\citet{lin2014coco}.

\subsection{Limitations}
\label{ssec:limitation}
While OLiVES addresses the challenges of other low-light video datasets, OLiVES has several limitations that we acknowledge: (1) the dataset is captured exclusively in outdoor daytime conditions with ND-filter-simulated low light rather than true low-light illumination; while this enables paired ground-truth capture, the noise and colour characteristics may differ from real low-light footage. (2) The scenes are restricted to a single geographic region, which may constrain generalisation. (3) The beam-splitter setup limits capture to tripod-based scenarios, excluding handheld and aerial footage. (4) While a subset of annotations is manually refined, the remaining training masks are generated by SAM2 without human verification and may contain errors.

\begin{table*}
    \centering
    \caption{Benchmarking on OLiVES dataset across recent LLVE and LLIE SOTA methods. \textbf{Bold} and \underline{underline} denote the best and second-best performances among all methods respectively.}
    \resizebox{1.0\linewidth}{!}{
    \begin{tabular}{lccccccccc}
    \toprule
    
    \textbf{Method} & \textbf{Venue} & \textbf{Type} & \textbf{Data} & \textbf{PSNR}$\uparrow$ & \textbf{SSIM}$\uparrow$ & \textbf{LPIPS}$\downarrow$ & \makecell{\textbf{tLPIPS}$\downarrow$\\$\times10^{-2}$} & \makecell{\textbf{WE}$\downarrow$\\$\times10^{-2}$} & \textbf{VMAF}$\uparrow$ \\
    
    \midrule
    SNR-aware~\cite{xu2022snr} & CVPR'22 & S & I & 26.20 & 0.835 & 0.268 & 6.47 & 5.97 & 73.51 \\
    LLFlow~\cite{wang2021low} & AAAI'22 & S & I & 21.65 & \underline{0.862} & \textbf{0.219} & 2.82 & \underline{3.81} & 69.50 \\
    Retinexformer~\cite{Cai2023retinexformer} & ICCV'23 & S & I & \textbf{27.65} & 0.834 & 0.241 & 6.35 & 5.41 & \textbf{76.93} \\
    LightenDiffusion~\cite{Jiang2024ECCVLightenDiff} & ECCV'24 & UP & I & 20.32 & 0.741 & 0.380 & 8.70 & 5.56 & 54.36 \\
    Wakeup-darkness~\cite{zhang2025wakeup} & ACM TOMM'25 & US & I & 17.46 & 0.534 & 0.630 & 8.82 & 9.26 & 71.39 \\
    
    \midrule
    SGZSL~\cite{Zheng2022SGZSL} & WACV'22 & US & V & 16.61 & 0.527 & 0.693 & 10.70 & 11.80 & \underline{74.74} \\
    Zero-TIG~\cite{li2025zerotig} & EUSIPCO'25 & US & V & 19.25 & 0.767 & 0.392 & 4.66 & 6.06 & 63.88 \\ 
    TempRetinex~\cite{li2026TempRetinex} & ICME'26 & US & V & 19.66 & 0.723 & 0.402 & 5.21 & \textbf{2.68} & 62.63 \\
    
    \midrule
    SDSD-Net~\cite{wang2021sdsd} & ICCV'21 & S & V & 22.59 & 0.716 & 0.322 & 6.00 & 7.58 & 65.15  \\
    StableLLVE~\cite{Zhang2021StableLLVE} & CVPR'21 & S & V & 22.60 & 0.851 & 0.300 & 2.80 & 4.48 & 49.35 \\
    DP3DF~\cite{xu2023dp3df} & AAAI'23 & S & V & 23.22 & 0.781 & 0.325 & 6.22 & 6.57 & 37.37 \\
    FastLLVE~\cite{li2023fastllve} & ACM MM'23 & S & V & 25.87 & 0.800 & 0.309 & 4.97 & 6.88 & 73.01 \\
    BVI-CDM~\cite{Lin2024BVICDM} & CVMP'24 & S & V & 26.40 & 0.854 & 0.326 & \textbf{1.67} & 3.90 & 56.03 \\
    BVI-Mamba~\cite{huang2025BVIMamba} & MLCD'25 & S & V & 25.66 & \textbf{0.884} & \underline{0.228} & \underline{1.92} & 3.97 & 71.14 \\
    LLVE-STCD~\cite{xu2025llvestcd} & IJCAI'25 & S & V & \underline{27.00} & 0.827 & 0.255 & 5.75 & 5.34 & 74.70\\

    \bottomrule
    \multicolumn{10}{l}{\scriptsize S: Supervised, US: Unsupervised, UP: Unpaired. I: Image, V: Video. WE: Warp Error.}
    \end{tabular}}
    \label{tab:llve_benchmark}
    \vspace{-2pt}
\end{table*}

\begin{table}
    \centering
    \caption{Benchmark of SOTA VOS methods trained on the OLiVES dataset and evaluated on OLiVES validation and test set, and LLE-VOS validation set. \textbf{Bold} and \underline{underline} denote the best and second-best performances among all methods respectively.}
    \resizebox{1.0\linewidth}{!}{
    \begin{tabular}{lcccccccccccc}
    \toprule
    \multirow{2}{*}{\textbf{Method}} & \multirow{2}{*}{\textbf{Venue}} & \multicolumn{4}{c}{\textbf{OLiVES val}} & \multicolumn{4}{c}{\textbf{OLiVES test}} & \multicolumn{3}{c}{\textbf{LLE-VOS~\cite{Li2024LLEVOS} val}} \\
     & &
     $\mathcal{J}\&\mathcal{F}$ & $\mathcal{J}$ & $\mathcal{F}$ & $\mathcal{B}$ 
     & $\mathcal{J}\&\mathcal{F}$ & $\mathcal{J}$ & $\mathcal{F}$  & $\mathcal{B}$ 
     & $\mathcal{J}\&\mathcal{F}$ & $\mathcal{J}$ & $\mathcal{F}$ \\
    \midrule
    STCN~\cite{cheng2021STCN} & NeurIPS'21 & 89.4 & 87.2 & 91.5 & 83.5 & 78.3 & 73.8 & 82.9 & 70.8 & 32.3 & 37.9 & 26.7 \\
    AOT~\cite{yang2021AOT} & NeurIPS'21 & 91.5 & 88.5 & 94.6 & 85.1 & 80.5 & 75.8 & 85.1 & 73.4 & 54.1 & 57.5 & 50.7 \\
    De-AOT~\cite{yang2022DeAOT} & NeurIPS'22 & 91.5 & 88.4 & 94.5 & 85.0 & 86.9 & 82.0 & 91.7 & 80.0 & \textbf{62.6} & \textbf{65.6} & \textbf{59.7} \\
    XMem~\cite{cheng2022xmem} & ECCV'22 & 86.9 & 84.4 & 89.5 & 79.5 & 74.4 & 68.9 & 79.8 & 65.3 & 53.0 & 56.0 & 50.1 \\
    Cutie~\cite{Cheng2024cutie} & CVPR'24 & \textbf{93.1} & \textbf{91.3} & \textbf{94.8} & \textbf{88.0} & \underline{88.3} & \textbf{84.2} & \underline{92.4} & \textbf{82.0} & \underline{59.2} & \underline{62.8} & 55.7 \\
    OASIS~\cite{qin2025oasis} & ICCV'25 & \underline{92.6} & \underline{90.8} & \underline{94.4} & \underline{87.7} & \textbf{88.4} & \underline{83.8} & \textbf{92.9} & \underline{81.7} & 59.1 & 62.5 & \underline{55.8} \\
    \bottomrule
    \end{tabular}}
    \label{tab:results_vos_methods}
    \vspace{-10pt}
\end{table}

\section{Experiments}
\label{sec:experiments}

\subsection{Implementation Details}
\label{ssec:details}
\paragraph{Compute resources.}
We conducted all of our experiments on a shared cluster where each node utilizes 4 NVIDIA GH200s superchips. One NVIDIA GH200 superchip is comprised of a 72-core NVIDIA Grace CPU (with 120GB CPU memory) and a NVIDIA H100 GPU (with 96GB GPU memory).
Each of the experiments we conducted varied greatly in runtime (due to different training configurations, architectures, etc.), but we estimate the total GPU hours is $\sim$500 hours.

\paragraph{LLVE training setup.}
We establish a comprehensive benchmark on OLiVES using a wide range of state-of-the-art (SOTA) methods for both LLIE and LLVE. For a fair comparison, supervised methods~\cite{xu2022snr,wang2021low,Cai2023retinexformer,wang2021sdsd,Zhang2021StableLLVE,xu2023dp3df,li2023fastllve,Lin2024BVICDM,huang2025BVIMamba,xu2025llvestcd}, and unsupervised methods~\cite{zhang2025wakeup,Zheng2022SGZSL,li2025zerotig,li2026TempRetinex} are trained on OLiVES with the settings from the original works. Notably, GT is incorporated as input priors in~\cite{Jiang2024ECCVLightenDiff} while excluded from the loss function during training; thus, we categorize it as an unpaired method.

\paragraph{VOS training setup.}
We employed several state-of-the-art semi-supervised VOS methods for our low-light VOS task, including STCN~\cite{cheng2021STCN}, AOT~\cite{yang2021AOT}, De-AOT~\cite{yang2022DeAOT}, XMem~\cite{cheng2022xmem}, and Cutie~\cite{Cheng2024cutie}. For fair comparison, we considered only architectures using a ResNet-50~\cite{he2016resnet} feature extractor. All methods were initialized using the authors’ publicly released weights pretrained on DAVIS 2017~\cite{PontTuset2017davis17} and YouTube-VOS~\cite{xu2018youtubevos18}, and then finetuned on our training set for 30,000 iterations using a batch size of 8 and a learning rate of 1$\times$10$^{-4}$. The only exception was OASIS~\cite{qin2025oasis}, for which only ResNet-18 pre-trained weights were publicly available. To ensure consistency, we first re-trained OASIS following its original training protocol on DAVIS 2017 and YouTube-VOS, before finetuning it on our dataset using the same settings as the other methods.

\subsection{Low-Light Video Enhancement}
For the evaluation metrics, we adopt PSNR, SSIM, LPIPS~\cite{zhang2018perceptual} to measure the image restoration quality. To further assess video quality and temporal consistency, we employ tLPIPS~\cite{chu2020tlpips}, Warping Error (WE)~\cite{Lai2018WE}, and VMAF~\cite{li2016vmaf}. The evaluate results on our test set is reported in~\cref{tab:llve_benchmark}. 

We observe a clear performance gap between supervised and unsupervised approaches, and between image-based and video-based methods:
(i) Image-based methods generally achieve higher PSNR and VMAF scores than video-based methods (e.g., Retinexformer~\cite{Cai2023retinexformer}), but often exhibit weaker temporal consistency, as reflected by higher tLPIPS and WE values. In contrast, video-based methods produce more temporally stable results; for example, BVI-CDM~\cite{Lin2024BVICDM} and BVI-Mamba~\cite{huang2025BVIMamba} achieve lower tLPIPS and competitive WE scores, albeit at the cost of reduced contrast. 
(ii) Unsupervised and unpaired methods exhibit a substantial performance drop compared to supervised approaches across all metrics. For example,~\cite{Zheng2022SGZSL} and~\cite{zhang2025wakeup} achieve lower PSNR and higher LPIPS, highlighting the difficulty of learning robust enhancement without paired supervision on our dataset.

Overall, methods trained on OLiVES struggle under severe degradations. While some approaches perform well on individual metrics, no single method consistently achieves strong results across both quality and temporal measures. This indicates the value of OLiVES as a challenging benchmark and impresses the need for joint optimization of restoration quality and temporal consistency. Further qualitative analysis is provided in~\cref{sec:suppmat_llve_qualitative}.

\subsection{Video Object Segmentation}
We provide a quantitative evaluation of several state-of-the-art VOS methods trained on our dataset, and assess their performance on our validation and test sets, as well as the LLE-VOS~\cite{Li2024LLEVOS} validation set, to demonstrate the generalization of our training data across low-light benchmarks.
We use standard VOS evaluation metrics~\cite{Perazzi2016davis16, PontTuset2017davis17}: Jaccard Index ($\mathcal{J}$), F-measure ($\mathcal{F}$) and $\mathcal{J}\&\mathcal{F}$. Results are reported in~\cref{tab:results_vos_methods}.
We report that~\cite{Cheng2024cutie} and~\cite{qin2025oasis} show the strongest performance on OLiVES and LLE-VOS, while~\cite{yang2022DeAOT} achieves the second-strongest performance on LLE-VOS. STCN~\cite{cheng2021STCN} performs poorly on LLE-VOS, suggesting that its earlier architecture struggles to generalize across domains.

\begin{wraptable}{r}{7cm}
    \centering
    \vspace{-18pt}
    \caption{Quantitative results of two-stage methods on LLE-VOS validation set, using LLVE methods trained on OLiVES and pre-trained Cutie~\cite{Cheng2024cutie}.}
    \resizebox{1.0\linewidth}{!}{
    \begin{tabular}{lccc}
    \toprule
    \multirow{2}{*}{\textbf{LLVE Method}} & \multicolumn{3}{c}{\textbf{LLE-VOS~\cite{Li2024LLEVOS} val}} \\
     & $\mathcal{J}\&\mathcal{F}$ & $\mathcal{J}$ & $\mathcal{F}$ \\
    \midrule
    Cutie (off-the-shelf)~\cite{Cheng2024cutie} & 53.2 & 58.5 & 48.0 \\
    LLVE-STCD~\cite{xu2025llvestcd} + Cutie & 57.9 & \textbf{62.1} & 53.8 \\
    BVI-Mamba~\cite{huang2025BVIMamba} + Cutie & 58.1 & 62.0 & 53.8 \\
    RetinexFormer~\cite{Cai2023retinexformer} + Cutie & \textbf{58.3} & 61.9 & \textbf{54.6} \\
    \bottomrule
    \end{tabular}}
    \label{tab:results_twostage_vos}
    \vspace{-10pt}
\end{wraptable}
We further evaluate the VOS performance of four LLVE methods~\cite{Cai2023retinexformer, huang2025BVIMamba, xu2025llvestcd} trained on OLiVES within a two-stage pipeline, where enhancement is applied as a pre-processing step before inference with a pre-trained VOS model. For this experiment, we use Cutie with off-the-shelf weights for segmentation after enhancement. Evaluation on the LLE-VOS validation set assesses the effectiveness of our training data for LLVE; results are reported in~\cref{tab:results_twostage_vos}.

We find that all VOS methods perform well even without fine-tuning on our dataset (see~\cref{sec:suppmat_vos_extra_experiments}), indicating strong object detectability. However, segmentation quality remains limited, with inaccurate boundaries and missing fine details (e.g., animal fur). This suggests that, while current methods are suitable for detection-oriented applications such as surveillance, they fall short for tasks requiring fine-grained detail, such as film post-production, augmented reality, and robotics manipulation, where accurate boundaries and fine structures are critical. We further analyze this behavior in~\cref{sec:suppmat_vos_high_performances}.

\section{Conclusion}
We present OLiVES, a new low-light video dataset supporting both low-light video enhancement (LLVE) and video object segmentation (VOS). OLiVES introduces diverse challenges, including complex motion, rich natural textures (e.g., textures of animal furs), and a wide range of illumination conditions that are widely absent by existing low-light datasets. In particular, it provides new data for low-light VOS, featuring occlusions, emerging objects, and degraded appearances due to motion blur, without reliance on auxiliary modalities such as event data.
By establishing a unified benchmark for both LLVE and VOS, OLiVES facilitates research that jointly advances perceptual quality and semantic understanding. We hope this dataset will encourage further work in low-light video analysis and inspire more robust methods for challenging real-world conditions.


{
\small
\bibliographystyle{plainnat} 
\bibliography{refs}

@inproceedings{wang2021sdsd,
  title={Seeing Dynamic Scene in the Dark: High-Quality Video Dataset with Mechatronic Alignment},
  author={Wang, Ruixing and Xu, Xiaogang and Fu, Chi-Wing and Lu, Jiangbo and Yu, Bei and Jia, Jiaya},
  booktitle={ICCV},
  year={2021}
}

@InProceedings{Fu2023DID,
    author    = {Fu, Huiyuan and Zheng, Wenkai and Wang, Xicong and Wang, Jiaxuan and Zhang, Heng and Ma, Huadong},
    title     = {Dancing in the Dark: A Benchmark towards General Low-light Video Enhancement},
    booktitle = {Proceedings of the IEEE/CVF International Conference on Computer Vision (ICCV)},
    month     = {October},
    year      = {2023},
    pages     = {12877-12886}
}

@article{lin2024BVIRLV,
  title={{BVI-RLV}: A Fully Registered Dataset and Benchmarks for Low-Light Video Enhancement},
  author={Lin, Ruirui and Anantrasirichai, Nantheera and Huang, Guoxi and Lin, Joanne and Sun, Qi and Malyugina, Alex and Bull, David},
  journal={arXiv preprint arXiv:2407.03535},
  year={2024}
}

@ARTICLE{chen2026evlight++,
  author={Chen, Kanghao and Liang, Guoqiang and Lu, Yunfan and Li, Hangyu and Wang, Lin},
  journal={IEEE Transactions on Pattern Analysis and Machine Intelligence}, 
  title={EvLight++: Low-Light Video Enhancement With an Event Camera: A Large-Scale Real-World Dataset, Novel Method, and More}, 
  year={2026},
  volume={48},
  number={2},
  pages={1608-1625},
  doi={10.1109/TPAMI.2025.3617801}}

@InProceedings{Li2024LLEVOS,
    author    = {Li, Hebei and Wang, Jin and Yuan, Jiahui and Li, Yue and Weng, Wenming and Peng, Yansong and Zhang, Yueyi and Xiong, Zhiwei and Sun, Xiaoyan},
    title     = {Event-assisted Low-Light Video Object Segmentation},
    booktitle = {Proceedings of the IEEE/CVF Conference on Computer Vision and Pattern Recognition (CVPR)},
    month     = {June},
    year      = {2024},
    pages     = {3250-3259}
}

@InProceedings{xu2018youtubevos18,
    author = {Xu, Ning and Yang, Linjie and Fan, Yuchen and Yang, Jianchao and Yue, Dingcheng and Liang, Yuchen and Price, Brian and Cohen, Scott and Huang, Thomas},
    title = {{YouTube-VOS}: Sequence-to-Sequence Video Object Segmentation},
    booktitle = {Proceedings of the European Conference on Computer Vision (ECCV)},
    month = {September},
    year = {2018}
}

@misc{yang2019youtubevos19,
  author       = "Linjie Yang and Yuchen Fan and Ning Xu",
  title        = "The 2nd Large-scale Video Object Segmentation Challenge - video object segmentation track",
  month        = oct,
  year         = "2019",
}

@article{PontTuset2017davis17,
  author = {Jordi Pont-Tuset and Federico Perazzi and Sergi Caelles and Pablo Arbel\'aez and Alexander Sorkine-Hornung and Luc {Van Gool}},
  title = {The 2017 {DAVIS} Challenge on Video Object Segmentation},
  journal = {arXiv:1704.00675},
  year = {2017}
}

@inproceedings{Perazzi2016davis16,
  author = {F. Perazzi and J. Pont-Tuset and B. McWilliams and L. {Van Gool} and M. Gross and A. Sorkine-Hornung},
  title = {A Benchmark Dataset and Evaluation Methodology for Video Object Segmentation},
  booktitle = {Computer Vision and Pattern Recognition},
  year = {2016}
}

@article{Ding2025MOSEv2,
    title={{MOSEv2}: A More Challenging Dataset for Video Object Segmentation in Complex Scenes},
    author={Ding, Henghui and Ying, Kaining and Liu, Chang and He, Shuting and Jiang, Xudong and Jiang, Yu-Gang and Torr, Philip HS and Bai, Song},
    journal={arXiv preprint arXiv:2508.05630},
    year={2025}
}

@inproceedings{Ding2023MOSE,
  title={{MOSE}: A New Dataset for Video Object Segmentation in Complex Scenes},
  author={Ding, Henghui and Liu, Chang and He, Shuting and Jiang, Xudong and Torr, Philip HS and Bai, Song},
  booktitle={ICCV},
  year={2023}
}

@inproceedings{cheng2021STCN,
    author = {Cheng, Ho Kei and Tai, Yu-Wing and Tang, Chi-Keung},
    title = {Rethinking space-time networks with improved memory coverage for efficient video object segmentation},
    year = {2021},
    isbn = {9781713845393},
    booktitle = {Proceedings of the 35th International Conference on Neural Information Processing Systems},
    numpages = {14},
}

@inproceedings{yang2021AOT,
    author = {Yang, Zongxin and Wei, Yunchao and Yang, Yi},
    title = {Associating objects with transformers for video object segmentation},
    year = {2021},
    isbn = {9781713845393},
    publisher = {Curran Associates Inc.},
    address = {Red Hook, NY, USA},
    booktitle = {Proceedings of the 35th International Conference on Neural Information Processing Systems},
    articleno = {191},
    numpages = {12},
    series = {NIPS '21}
}

@inproceedings{yang2022DeAOT,
    author = {Yang, Zongxin and Yang, Yi},
    title = {Decoupling features in hierarchical propagation for video object segmentation},
    year = {2022},
    isbn = {9781713871088},
    publisher = {Curran Associates Inc.},
    address = {Red Hook, NY, USA},
    booktitle = {Proceedings of the 36th International Conference on Neural Information Processing Systems},
    articleno = {2632},
    numpages = {13},
    location = {New Orleans, LA, USA},
    series = {NIPS '22}
}

@InProceedings{cheng2022xmem,
    author="Cheng, Ho Kei
    and Schwing, Alexander G.",
    editor="Avidan, Shai
    and Brostow, Gabriel
    and Ciss{\'e}, Moustapha
    and Farinella, Giovanni Maria
    and Hassner, Tal",
    title="XMem: Long-Term Video Object Segmentation with an Atkinson-Shiffrin Memory Model",
    booktitle="Computer Vision -- ECCV 2022",
    year="2022",
    publisher="Springer Nature Switzerland",
    address="Cham",
    pages="640--658",
    isbn="978-3-031-19815-1"
}

@InProceedings{Cheng2024cutie,
    author    = {Cheng, Ho Kei and Oh, Seoung Wug and Price, Brian and Lee, Joon-Young and Schwing, Alexander},
    title     = {Putting the Object Back into Video Object Segmentation},
    booktitle = {Proceedings of the IEEE/CVF Conference on Computer Vision and Pattern Recognition (CVPR)},
    month     = {June},
    year      = {2024},
    pages     = {3151-3161}
}

@InProceedings{qin2025oasis,
    author    = {Qin, Guanyi and Wang, Ziyue and Shen, Daiyun and Liu, Haofeng and Zhou, Hantao and Wu, Junde and Hu, Runze and Jin, Yueming},
    title     = {Structure Matters: Revisiting Boundary Refinement in Video Object Segmentation},
    booktitle = {Proceedings of the IEEE/CVF International Conference on Computer Vision (ICCV)},
    month     = {October},
    year      = {2025},
    pages     = {14431-14442}
}

@InProceedings{lin2026elvis,
    author={Lin, Joanne and Lin, Ruirui and Li, Yini and Bull, David and Anantrasirichai, Nantheera},
    title={{ELVIS}: Enhance Low-Light for Video Instance Segmentation in the Dark},
    booktitle = {Proceedings of the IEEE/CVF Conference on Computer Vision and Pattern Recognition (CVPR)},
    month     = {June},
    year      = {2026}
}

@article{chen2023lis,
    title={Instance Segmentation in the Dark},
    author={Chen, Linwei and Fu, Ying and Wei, Kaixuan and Zheng, Dezhi and Heide, Felix},
    journal={International Journal of Computer Vision},
    volume={131},
    number={8},
    pages={2198--2218},
    year={2023},
    publisher={Springer}
}

@INPROCEEDINGS{lin2025wnlb,
  author={Lin, Joanne and Anantrasirichai, Nantheera and Bull, David},
  booktitle={ICASSP 2025 - 2025 IEEE International Conference on Acoustics, Speech and Signal Processing (ICASSP)}, 
  title={Multi-Scale Denoising in the Feature Space for Low-Light Instance Segmentation}, 
  year={2025},
  volume={},
  number={},
  pages={1-5},
  doi={10.1109/ICASSP49660.2025.10889336}
}

@inproceedings{guo2020zerodce,
 author = {Guo, Chunle Guo and Li, Chongyi and Guo, Jichang and Loy, Chen Change and Hou, Junhui and Kwong, Sam and Cong, Runmin},
 title = {Zero-reference deep curve estimation for low-light image enhancement},
 booktitle = {Proceedings of the IEEE conference on computer vision and pattern recognition (CVPR)},
 pages    = {1780-1789},
 month = {June},
 year = {2020}
}

@inproceedings{
    ravi2025sam2,
    title={{SAM} 2: Segment Anything in Images and Videos},
    author={Nikhila Ravi and Valentin Gabeur and Yuan-Ting Hu and Ronghang Hu and Chaitanya Ryali and Tengyu Ma and Haitham Khedr and Roman R{\"a}dle and Chloe Rolland and Laura Gustafson and Eric Mintun and Junting Pan and Kalyan Vasudev Alwala and Nicolas Carion and Chao-Yuan Wu and Ross Girshick and Piotr Dollar and Christoph Feichtenhofer},
    booktitle={The Thirteenth International Conference on Learning Representations},
    year={2025},
    url={https://openreview.net/forum?id=Ha6RTeWMd0}
}

@INPROCEEDINGS{he2016resnet,
  author={He, Kaiming and Zhang, Xiangyu and Ren, Shaoqing and Sun, Jian},
  booktitle={2016 IEEE Conference on Computer Vision and Pattern Recognition (CVPR)}, 
  title={Deep Residual Learning for Image Recognition}, 
  year={2016},
  volume={},
  number={},
  pages={770-778},
  doi={10.1109/CVPR.2016.90}}

@InProceedings{lin2014coco,
    author="Lin, Tsung-Yi
    and Maire, Michael
    and Belongie, Serge
    and Hays, James
    and Perona, Pietro
    and Ramanan, Deva
    and Doll{\'a}r, Piotr
    and Zitnick, C. Lawrence",
    editor="Fleet, David
    and Pajdla, Tomas
    and Schiele, Bernt
    and Tuytelaars, Tinne",
    title="Microsoft COCO: Common Objects in Context",
    booktitle="Computer Vision -- ECCV 2014",
    year="2014",
    publisher="Springer International Publishing",
    address="Cham",
    pages="740--755",
    isbn="978-3-319-10602-1"
}

@InProceedings{edstedt2024roma,
    author    = {Edstedt, Johan and Sun, Qiyu and B\"okman, Georg and Wadenb\"ack, M\r{a}rten and Felsberg, Michael},
    title     = {{RoMa}: Robust Dense Feature Matching},
    booktitle = {Proceedings of the IEEE/CVF Conference on Computer Vision and Pattern Recognition (CVPR)},
    month     = {June},
    year      = {2024},
    pages     = {19790-19800}
}

@inproceedings{chen2019seeing,
  title={Seeing motion in the dark},
  author={Chen, Chen and Chen, Qifeng and Do, Minh N and Koltun, Vladlen},
  booktitle={Proceedings of the IEEE/CVF International conference on computer vision},
  pages={3185--3194},
  year={2019}
}

@ARTICLE{Fu2022LLRVD,
  author={Fu, Ying and Wang, Zichun and Zhang, Tao and Zhang, Jun},
  journal={IEEE Transactions on Multimedia}, 
  title={Low-light Raw Video Denoising with a High-quality Realistic Motion Dataset}, 
  year={2022},
  volume={},
  number={},
  pages={1-13},
  doi={10.1109/TMM.2022.3233247}}

@INPROCEEDINGS{Li2018MegaDepth,
  author={Li, Zhengqi and Snavely, Noah},
  booktitle={2018 IEEE/CVF Conference on Computer Vision and Pattern Recognition}, 
  title={MegaDepth: Learning Single-View Depth Prediction from Internet Photos}, 
  year={2018},
  volume={},
  number={},
  pages={2041-2050},
  doi={10.1109/CVPR.2018.00218}}

@inproceedings{he2020momentum,
  title={Momentum contrast for unsupervised visual representation learning},
  author={He, Kaiming and Fan, Haoqi and Wu, Yuxin and Xie, Saining and Girshick, Ross},
  booktitle={Proceedings of the IEEE/CVF conference on computer vision and pattern recognition},
  pages={9729--9738},
  year={2020}
}

@inproceedings{bardes2022vicreg,
  author  = {Adrien Bardes and Jean Ponce and Yann LeCun},
  title   = {VICReg: Variance-Invariance-Covariance Regularization For Self-Supervised Learning},
  booktitle = {ICLR},
  year    = {2022},
}

@inproceedings{li2026TempRetinex,
      title={TempRetinex: Retinex-based Unsupervised Enhancement for Low-light Video Under Diverse Lighting Conditions}, 
      author={Yini Li and Louis Forster and David Bull and Nantheera Anantrasirichai},
      year={2026},
      booktitle = {2026 ICME},
      url={https://arxiv.org/abs/2511.09609}, 
}

@InProceedings{Zhang2021StableLLVE,
    author    = {Zhang, Fan and Li, Yu and You, Shaodi and Fu, Ying},
    title     = {Learning Temporal Consistency for Low Light Video Enhancement From Single Images},
    booktitle = {Proceedings of the IEEE/CVF Conference on Computer Vision and Pattern Recognition (CVPR)},
    month     = {June},
    year      = {2021},
    pages     = {4967-4976}
}

@inproceedings{xu2023dp3df,
  title={Deep Parametric 3D Filters for Joint Video Denoising and Illumination Enhancement in Video Super Resolution},
  author={Xu, Xiaogang and Wang, Ruixing and Fu, Chi-Wing and Jia, Jiaya},
  booktitle={AAAI},
  year={2023}
}

@InProceedings{li2023fastllve,
  title={FastLLVE: Real-Time Low-Light Video Enhancement with Intensity-Aware Lookup Table},
  author={Li, Wenhao and Wu, Guangyang and Wang, Wenyi and Ren, Peiran and Liu, Xiaohong},
  booktitle={Proceedings of the 31th ACM International Conference on Multimedia},
  year={2023}
}

@inproceedings{Lin2024BVICDM,
author = {Lin, Ruirui and Sun, Qi and Anantrasirichai, Nantheera},
title = {Low-light Video Enhancement with Conditional Diffusion Models and Wavelet Interscale Attentions},
year = {2024},
booktitle = {Proceedings of 21st ACM SIGGRAPH Conference on Visual Media Production},
}

@inproceedings{huang2025BVIMamba,
title = "BVI-Mamba: Video Enhancement Using a Visual State-Space Model for Low-Light and Underwater Environments",
author = "G. Huang and R. Lin and Y. Li and D. Bull and N. Anantrasirichai",
year = "2025",
booktitle = "Machine Learning from Challenging Data 2025",
}

@inproceedings{xu2025llvestcd,
  title={Low-Light Video Enhancement via Spatial-Temporal Consistent Decomposition},
  author={Xu, Xiaogang and Zhou, Kun and Hu, Tao and Wu, Jiafei and Wang, Ruixing and Peng, Hao and Yu, Bei},
  booktitle={IJCAI},
  year={2025}
}

@inproceedings{li2025zerotig,
  title={Zero-TIG: Temporal Consistency-Aware Zero-Shot Illumination-Guided Low-light Video Enhancement},
  author={Y. Li and N. Anantrasirichai},
  booktitle={EUSIPCO},
  year={2025}
}

@inproceedings{Hill2026DMAT,
  title={{DMAT}: An End-to-End Framework for Joint Atmospheric Turbulence Mitigation and Object Detection},
  author={Paul Hill and Zhiming Liu and Alin Achim and Dave Bull and Nantheera Anantrasirichai},
  booktitle={IEEE/CVF WACV},
  pages={2690--2699},
  year={2026}
}

@inproceedings{Zheng2022SGZSL,
  title={Semantic-guided zero-shot learning for low-light image/video enhancement},
  author={S. Zheng and G. Gupta},
  booktitle={IEEE/CVF WACVW},
  pages={581--590},
  year={2022}
}

@article{zhang2025wakeup,
  title={Wakeup-Darkness: When Multimodal Meets Unsupervised Low-light Image Enhancement},
  author={Zhang, Xiaofeng and Xu, Zishan and Tang, Hao and Gu, Chaochen and Chen, Wei},
  journal={ACM Transactions on Multimedia Computing, Communications and Applications},
  year={2025}
}

@InProceedings{Jiang2024ECCVLightenDiff,
    author    = {Jiang, Hai and Luo, Ao and Liu, Xiaohong and Han, Songchen and Liu, Shuaicheng},
    title     = {LightenDiffusion: Unsupervised Low-Light Image Enhancement with Latent-Retinex Diffusion Models},
    booktitle = {European Conference on Computer Vision},
    year      = {2024},

}

@inproceedings{Cai2023retinexformer,
  title={Retinexformer: One-stage Retinex-based Transformer for Low-light Image Enhancement},
  author={Cai, Yuanhao and Bian, Hao and Lin, Jing and Wang, Haoqing and Timofte, Radu and Zhang, Yulun},
  booktitle={ICCV},
  year={2023}
}

@inproceedings{xu2022snr,
  title={SNR-aware Low-Light Image Enhancement},
  author={Xu, Xiaogang and Wang, Ruixing and Fu, Chi-Wing and Jia, Jiaya},
  booktitle={CVPR},
  year={2022}
}

@inproceedings{wang2021low,
  title={Low-Light Image Enhancement with Normalizing Flow},
  author={Wang, Yufei and Wan, Renjie and Yang, Wenhan and Li, Haoliang and Chau, Lap-Pui and Kot, Alex C},
  booktitle={AAAI Conference on Artificial Intelligence},
  year={2022}
}

@inproceedings{Lai2018WE,
    author    = {Lai, Wei-Sheng and Huang, Jia-Bin and Wang, Oliver and Shechtman, Eli and Yumer, Ersin and Yang, Ming-Hsuan}, 
    title     = {Learning Blind Video Temporal Consistency}, 
    booktitle = {European Conference on Computer Vision},
    year      = {2018}
}

@InProceedings{zhang2018perceptual,
author = {Zhang, Richard and Isola, Phillip and Efros, Alexei A. and Shechtman, Eli and Wang, Oliver},
title = {The Unreasonable Effectiveness of Deep Features as a Perceptual Metric},
booktitle = {Proceedings of the IEEE Conference on Computer Vision and Pattern Recognition (CVPR)},
month = {June},
year = {2018}
}

@article{chu2020tlpips,
  title={Learning temporal coherence via self-supervision for GAN-based video generation},
  author={Chu, Mengyu and Xie, You and Mayer, Jonas and Leal-Taix{\'e}, Laura and Thuerey, Nils},
  journal={ACM Transactions on Graphics (TOG)},
  volume={39},
  number={4},
  pages={75--1},
  year={2020},
  publisher={ACM New York, NY, USA}
}

@inproceedings{li2016vmaf,
  title={Toward A Practical Perceptual Video Quality Metric},
  author={Li, Zhi and Aaron, Anne and Katsavounidis, Ioannis and Moorthy, Anush Krishna and Manohara, Mehdi},
  booktitle={Netflix Tech Blog},
  volume={6},
  number={2},
  year={2016}
}

@inproceedings{liu2025jdatt,
  title     = {{JDATT}: A Joint Distillation Framework for Atmospheric Turbulence Mitigation and Target Detection},
  author    = {Liu, Zhiming and Hill, Paul and Anantrasirichai, Nantheera},
  booktitle = {Proceedings of the British Machine Vision Conference (BMVC)},
  year      = {2025}
}

@inproceedings{2022_1cac8326,
 author = {Sun, Shangquan and Ren, Wenqi and Wang, Tao and Cao, Xiaochun},
 booktitle = {Advances in Neural Information Processing Systems},
 editor = {S. Koyejo and S. Mohamed and A. Agarwal and D. Belgrave and K. Cho and A. Oh},
 pages = {4461--4474},
 publisher = {Curran Associates, Inc.},
 title = {Rethinking Image Restoration for Object Detection},
 url = {https://proceedings.neurips.cc/paper_files/paper/2022/file/1cac8326ce3fbe79171db9754211530c-Paper-Conference.pdf},
 volume = {35},
 year = {2022}
}

@InProceedings{Gu2025lowlightfoco,
    author    = {Gu, Yuxuan and Wang, Haoxuan and Ling, Pengyang and Wei, Zhixiang and Chen, Huaian and Jin, Yi and Chen, Enhong},
    title     = {Improving Visual and Downstream Performance of Low-Light Enhancer with Vision Foundation Models Collaboration},
    booktitle = {Proceedings of the IEEE/CVF Conference on Computer Vision and Pattern Recognition (CVPR)},
    month     = {June},
    year      = {2025},
    pages     = {16071-16080}
}
}


\appendix
\crefalias{section}{appendix}

\section{Details of RoMA finetuning}
\label{sec:suppmat_roma}

\paragraph{Synthetic low-light data.}
Following~\citet{lin2026elvis}, we model degradation of one clean normal-light image $I_{gt}$ as:
\begin{align}
I_{low}^{'} = {Deg}(I_{gt}, \phi) = H * (2^{\epsilon} I_{gt}) + N,
\label{eq:elvis}
\end{align}
where $*$ is the 2-D convolution, $\phi$ denotes degradation parameters including exposure adjustment $\epsilon$, a point spread function kernel $H$, and physics-based noise map $N$. Here we denote $\Phi$ as a set of $\phi$ estimated from multiple real low-light video frames using VDPNet~\cite{lin2026elvis}. Since different ND filters produce distinct brightness and noise distributions, we further construct degradation parameter sets:
\begin{align}
\Phi = \{\Phi_{0}, \Phi_{1.2}, \Phi_{1.5}, \Phi_{1.8}\},
\label{eq:phi}
\end{align}
while $\Phi_{0}$ corresponds to the identity mapping.

\paragraph{Degradation-augmented training.}
RoMa consists of a frozen coarse encoder $\mathcal{F}_{\text{coarse}}$ and a VGG19-based fine encoder $\mathcal{F}_{\text{fine}}$, where $\mathcal{F}_{\text{coarse}}$ extracts global correspondences and $\mathcal{F}_{\text{fine}}$ refines local matches. We observe that low-light degradation mainly affects $\mathcal{F}_{\text{fine}}$. To validate this, we conduct an ablation study on MegaDepth-1500~\cite{Li2018MegaDepth} using the pretrained RoMa model, where we replace either the coarse or fine features with those extracted from degraded images. As shown in~\cref{tab:roma_replace}, replacing fine features results in a substantial drop across all metrics. 

Therefore, we only finetune $\mathcal{F}_{\text{fine}}$ and the following refinement module $\mathcal{R}$.
During training, given an image pair $(I_A, I_B)$, we randomly apply degradation:
\begin{align}
I_{B}^{'} = {Deg}(I_{B}, \phi_1), \ \phi_1 \sim \{\Phi_{1.2}, \Phi_{1.5}, \Phi_{1.8}\},
\label{eq:deg1}
\end{align}
and use $(I_A, I_B^{'})$ as input for training. Following~\cite{edstedt2024roma}, we use MegaDepth~\cite{Li2018MegaDepth} as training set.

\begin{table}[htbp]
    \caption{Ablation study on the impact of coarse and fine features in RoMa. We replace the corresponding features in the pretrained model with those extracted from degraded inputs.}
    \centering
    \begin{tabular}{lccc}
    \toprule
    \textbf{Method} & \textbf{AUC@5} & \textbf{AUC@10} & \textbf{AUC@20} \\
    \midrule
    Pretrained &62.28 &76.45 &86.25\\
    \midrule
    Replace Coarse Features &62.19 &76.35 &86.15 \\
    Replace Fine Features &52.33 &68.68 &80.79\\
    \bottomrule
    \end{tabular}
    \label{tab:roma_replace}
\end{table}

\paragraph{Degradation-invariant feature learning.}
To further enhance robustness, we enforce degradation-invariant feature representations. Specifically, we generate an additional degraded sample and extract features as follow:
\begin{align}
I_{B}^{''} = {Deg}(I_{B}, \phi_2),\  \phi_2 \sim \Phi, \ \phi_2 \neq\phi_1, \\
\psi_{B}^{''}=\mathcal{F}_{\text{fine}}(I_{B}^{''}), \quad  \psi_{B}^{'}=\mathcal{F}_{\text{fine}}(I_{B}^{'}) .
\label{eq:deg2}
\end{align}
We encourage consistency between $\psi_B^{'}$ and $\psi_B^{''}$ using contrastive learning.
Inspired by VICReg~\cite{bardes2022vicreg}, we introduce an MLP predictor $\mathcal{P}$ to obtain projected feature embeddings in~\cref{eq:predictor}. We then apply the VICReg loss to enforce feature consistency and prevent representational collapse:
\begin{align}
\mathbf{z}_{B}^{'} = \mathcal{P}(\psi_{B}^{'}), \quad 
\mathbf{z}_{B}^{''} = \mathcal{P}(\psi_{B}^{''}).
\label{eq:predictor}
\end{align}
\begin{align}
\mathcal{L}_{\text{VICReg}} = 
\|\mathbf{z}_{B}^{'} - \mathbf{z}_{B}^{''}\|_2^2
+ \sum_{j} \max(0, \gamma - \sigma(\mathbf{z}_{B}^{'})_j)
+ \sum_{i \neq j} \text{Cov}(\mathbf{z}_{B}^{'})_{ij}^2,
\label{eq:VICReg}
\end{align}
where $\sigma(\mathbf{z})_j$ denotes the standard deviation of the $j$-th feature dimension over the batch, $\gamma$ denotes a minimum standard deviation, and $\text{Cov}(\mathbf{z})_{ij}$ denotes the $(i,j)$ entry of the covariance matrix of the projected features.

Following \cite{he2020momentum}, we adopt an InfoNCE to improve instance-level discriminability in a batch of size $K$:
\begin{align}
\mathcal{L}_{\text{InfoNCE}} = 
- \log \frac{\exp((z_{B}^{'T} \cdot z_{B}^{''}) / 0.07)}
{\sum^{K}_{k=0} \exp((z_{B}^{'T} \cdot z_{B,k}^{''}) / 0.07)},
\label{eq:InfoNCE}
\end{align}

Both InfoNCE and VICReg losses are added to the original model loss. Notably, $I_B^{''}$ is only used to improve feature robustness during training and not used for the final matching.

\paragraph{Results.}
All the training and testing experiments are conducted on a single NVIDIA H100 GPU on the compute platform described in~\cref{ssec:details}. Training duration is approximately 22 hours.
We analysis finetuned performances on MegaDepth-1500~\cite{Li2018MegaDepth} where the area under the cumulative curve (AUC) of Essential matrix error at thresholds ($5^\circ$, $10^\circ$, $20^\circ$) are reported, as shown in \cref{tab:RoMa_finetune}. Degradation augmentation (I) improves performance over all metrics. Adding VICReg (II) further boosts performance under degradation, while maintaining comparable results on non-degraded data. Incorporating InfoNCE (III) achieves the best results across all settings, improving both degraded and non-degraded performance. Finally, we select model III for our dataset registration. More implemented details and code can be referred to \url{https://github.com/anonymous-OLiVES/RoMa_Finetune4OLiVES}.

\begin{align}
\mathcal{L}_{\text{InfoNCE}} =
-\frac{1}{N}\sum_{i=1}^{N}
\log
\frac{\exp(\mathbf{z}_i^{\prime\top}\mathbf{z}_i^{\prime\prime}/\tau)}
{\sum_{j=1}^{N}\exp(\mathbf{z}_i^{\prime\top}\mathbf{z}_j^{\prime\prime}/\tau)}
\end{align}

\begin{table}[htbp]
    \caption{Finetuning performance of RoMa on non-degraded and degraded MegaDepth-1500. All values (except the first row) indicate absolute improvements over the pretrained baseline.}
    \centering
    \begin{tabular}{lcccccc}
    \toprule
    \multirow{2}{*}{\textbf{Method}} & \multicolumn{3}{c}{\textbf{Non-degraded}} & \multicolumn{3}{c}{\textbf{Degraded}} \\
    & \textbf{AUC@5} & \textbf{AUC@10} & \textbf{AUC@20} & \textbf{AUC@5} & \textbf{AUC@10} & \textbf{AUC@20} \\
    \midrule
    Pretrained &62.28 &76.45 &86.25 &54.75 &70.42 &82.04 \\
    \midrule
    I: Deg. Aug. &+0.34 &+0.11 &+0.04 &+2.43 &+2.06 &+1.46 \\
    II: I+VICReg &+0.10 &-0.03 &-0.06 &+2.74 &+2.17 &+1.49 \\
    III: II+InfoNCE &\textbf{+0.56} &\textbf{+0.38} &\textbf{+0.17} &\textbf{+3.12} &\textbf{+2.44} &\textbf{+1.66}\\
    \bottomrule
    \end{tabular}
    \label{tab:RoMa_finetune}
\end{table}

\begin{figure}
    \centering
    \includegraphics[width=\linewidth]{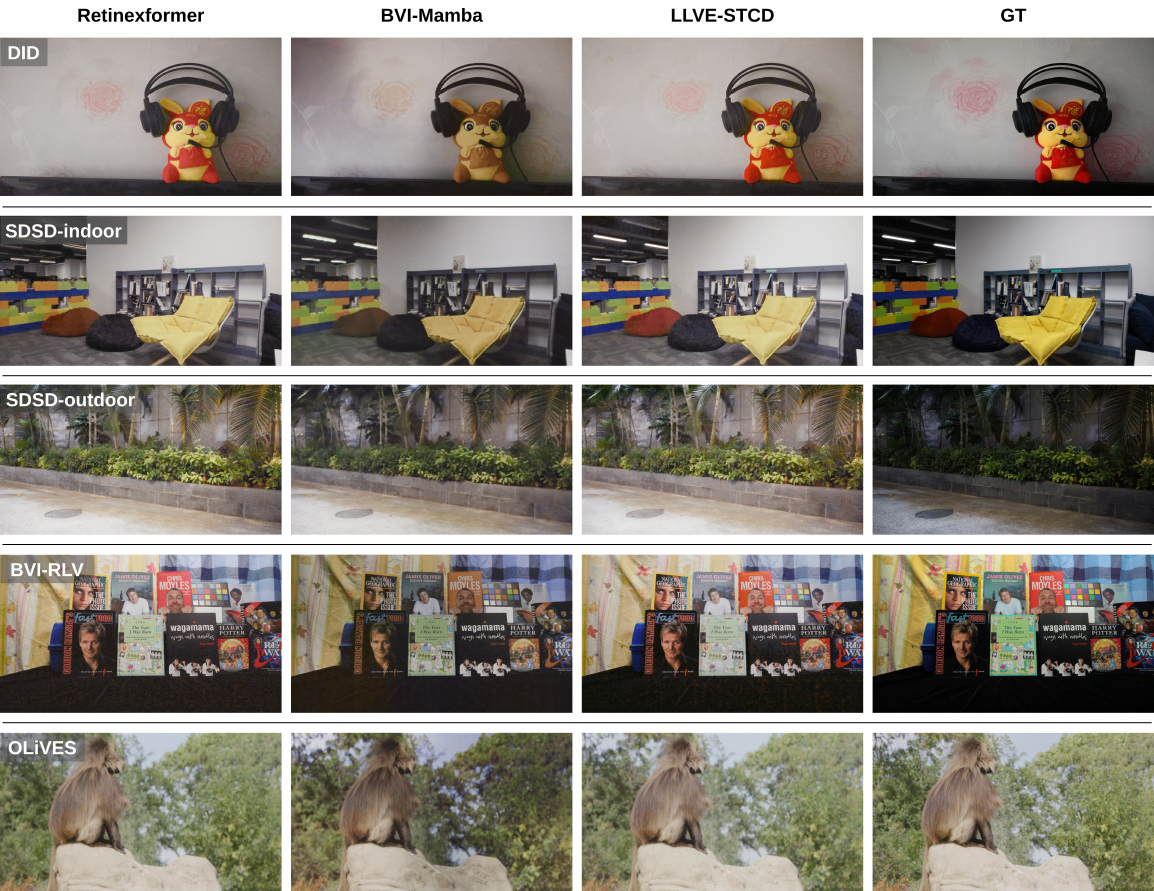}
    \caption{Qualitative results of LLVE methods trained on OLiVES.}
    \label{fig:results_visual_llve}
\end{figure}

\section{LLVE Qualitative Results}
\label{sec:suppmat_llve_qualitative}

We visualize the testing results of OLiVES on three representative SOTA methods (Retinexformer~\cite{Cai2023retinexformer}, BVI-Mamba~\cite{huang2025BVIMamba}, LLVE-STCD~\cite{xu2025llvestcd}).
To further evaluate the effectiveness of our dataset, we assess the cross-dataset generalization by testing on DID~\cite{Fu2023DID}, SDSD~\cite{wang2021sdsd}, BVI-RLV~\cite{lin2024BVIRLV}. As shown in Fig.~\ref{fig:results_visual_llve}, models trained on OLiVES produce visually pleasing results with improved brightness and detail restoration.
These results demonstrate that OLiVES enables models to learn more robust and transferable representations, leading to strong cross-dataset generalization and highlighting the effectiveness of our dataset.

\section{Details of VOS Annotation Process}
\label{sec:suppmat_vos_annotators}

\begin{figure}
    \centering
    \includegraphics[width=\linewidth]{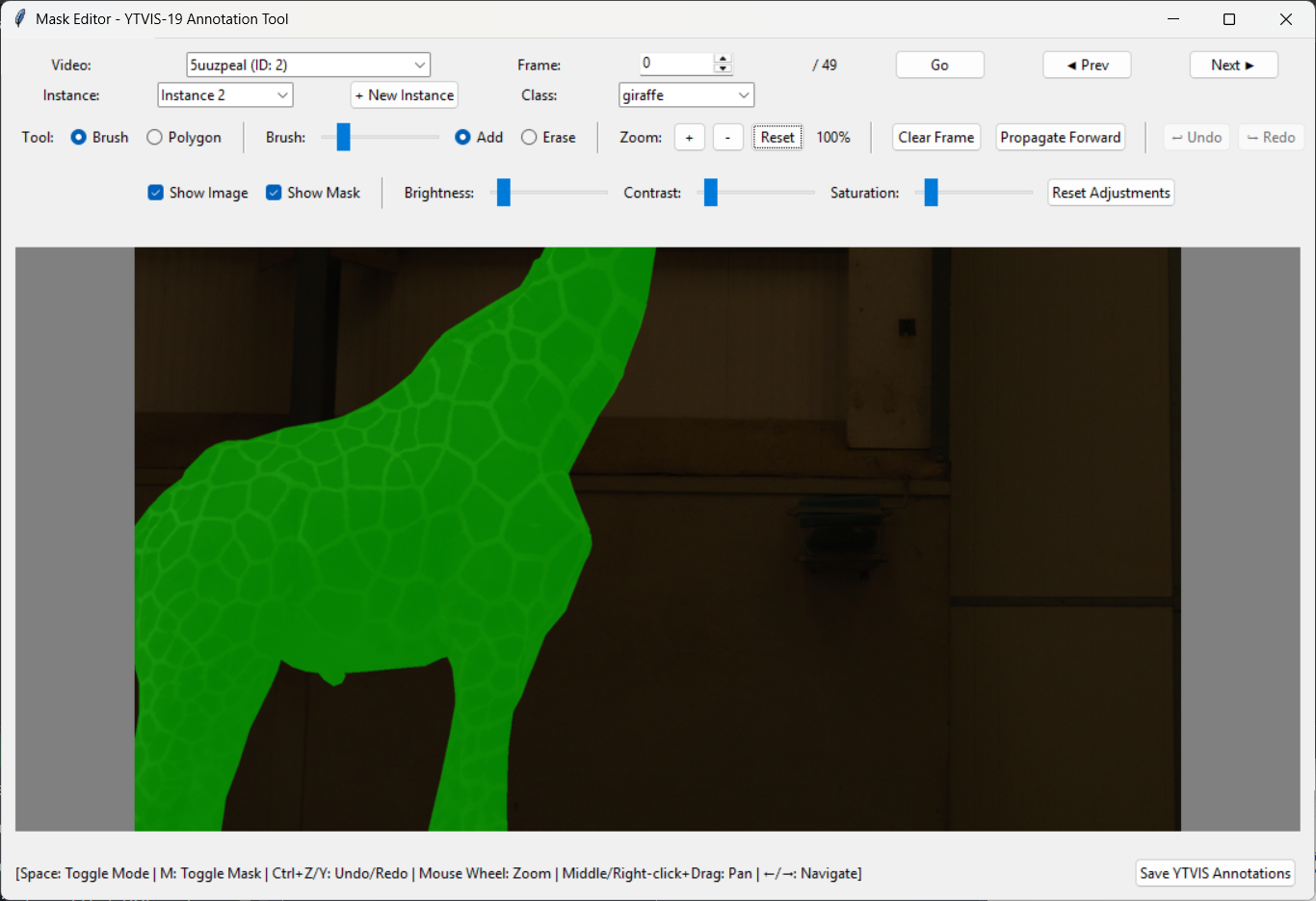}
    \caption{Custom annotation software to create manual annotations for OLiVES VOS benchmark.}
    \label{fig:annotation_tool}
\end{figure}

We recruited 9 annotators to manually produce ground-truth labels for the OLiVES test set by using our custom annotation software. We provided annotators with both the normal and low-light frames; the normal-light frames were for initial annotations and the low-light frames were to verify that they are correct. To ensure consistency and reduce ambiguity, annotators were provided with the first-frame object masks indicating when each instance appears in the sequence. We also provided them with custom annotation software, alongside detailed training materials (video tutorial and written instructions). The software included sliders to adjust brightness, contrast and saturation to help with verifying masks in low-light. We provide a screenshot of the tool in~\cref{fig:annotation_tool}.

Annotations were collected on a weekly basis, allowing us to perform regular quality assessments and provide targeted feedback. Annotators who consistently failed to meet quality standards were replaced, and their assigned videos were re-annotated by other workers. Following this initial stage, we introduced a second stage in which annotators reviewed and refined sequences they had not originally labeled, helping to improve consistency across the dataset.
Finally, we conducted an additional round of review to perform minor corrections and ensure a uniformly high standard of annotation quality. All annotators were compensated at rates above the local minimum wage.

\section{Further VOS Experiments}
\label{sec:suppmat_vos_extra_experiments}

\begin{figure}
    \centering
    \includegraphics[width=\linewidth]{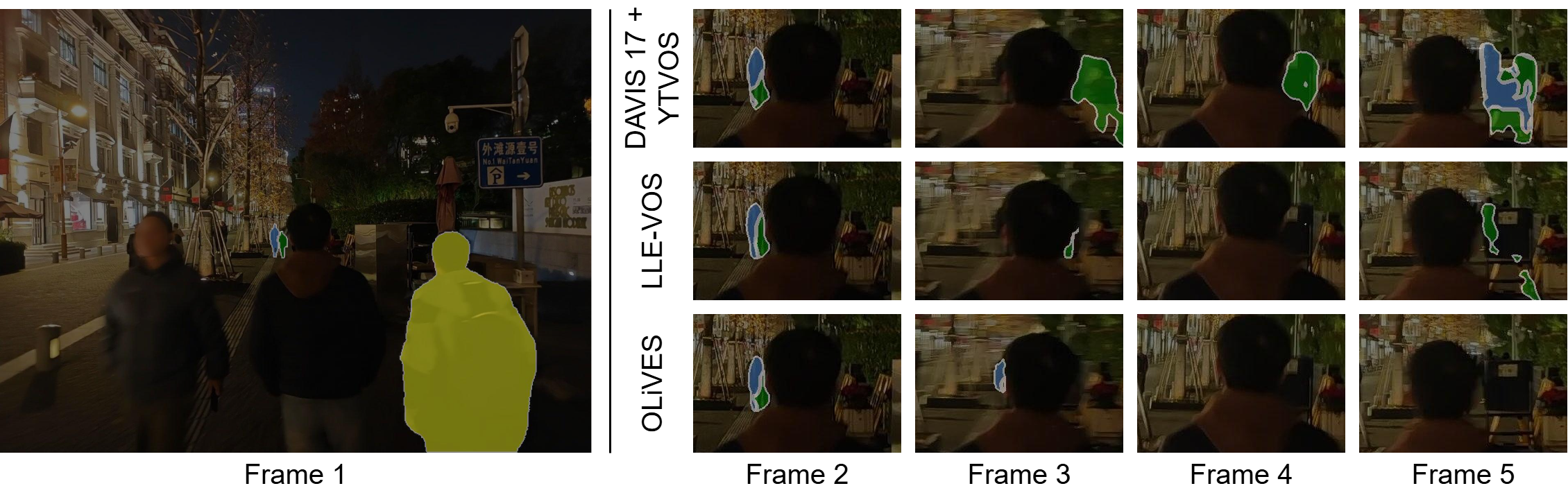}
    \caption{Qualitative comparison of Cutie on an example video from MOSEv2~\cite{Ding2025MOSEv2} (reduced frame brightness for better segmentation visibility). (Left) the first frame with the input masks. (Right) Crops of segmentation outputs for Cutie~\cite{Cheng2024cutie} with various training data.}
    \label{fig:results_vos_dataset_comparison}
\end{figure}

\paragraph{Cross-domain comparison.}Furthermore, to show the cross-domain capabilities of our training data for low-light VOS tasks, we compare Cutie~\cite{Cheng2024cutie} with its off-the-shelf weights (DAVIS 2017~\cite{PontTuset2017davis17} and YouTube-VOS~\cite{xu2018youtubevos18}), finetuned on LLE-VOS~\cite{Li2024LLEVOS} and finetuned on OLiVES.
We run inference on an example low-light video from the MOSEv2~\cite{Ding2025MOSEv2} dataset and visualize our results in~\cref{fig:results_vos_dataset_comparison}. Here, we find using the pre-trained weights, Cutie incorrectly produces segmentation masks for the people in the distance despite being occluded by a person in the foreground, while the finetuned model on LLE-VOS performed better with the occlusion but still produced masks in frame 5 despite the person not reappearing in the scene. Meanwhile, the model finetuned on OLiVES was able to correctly produce no segmentation masks when occluded.

\section{Investigation of High VOS performance}
\label{sec:suppmat_vos_high_performances}

\begin{table}
    \centering
    \caption{Quantitative results on OLiVES test set, grouped by size of the objects. Count is the total count of the object sizes.}
    \begin{tabular}{lcccc}
    \toprule
    \multirow{2}{*}{\textbf{Size}} & \multirow{2}{*}{\textbf{Count}} & \multicolumn{3}{c}{\textbf{OLiVES test}} \\
     & & $\mathcal{J}\&\mathcal{F}$ & $\mathcal{J}$ & $\mathcal{F}$ \\
    \midrule
     Small  &  18 & 75.0 & 65.1 & 84.9 \\
     Medium &  75 & 81.6 & 70.8 & 92.3 \\
     Large  & 151 & 88.5 & 86.2 & 90.7 \\
     \bottomrule
    \end{tabular}
    \label{tab:suppmat_olives_size_vos_scores}
\end{table}

\begin{figure}
    \centering
    \includegraphics[width=\linewidth]{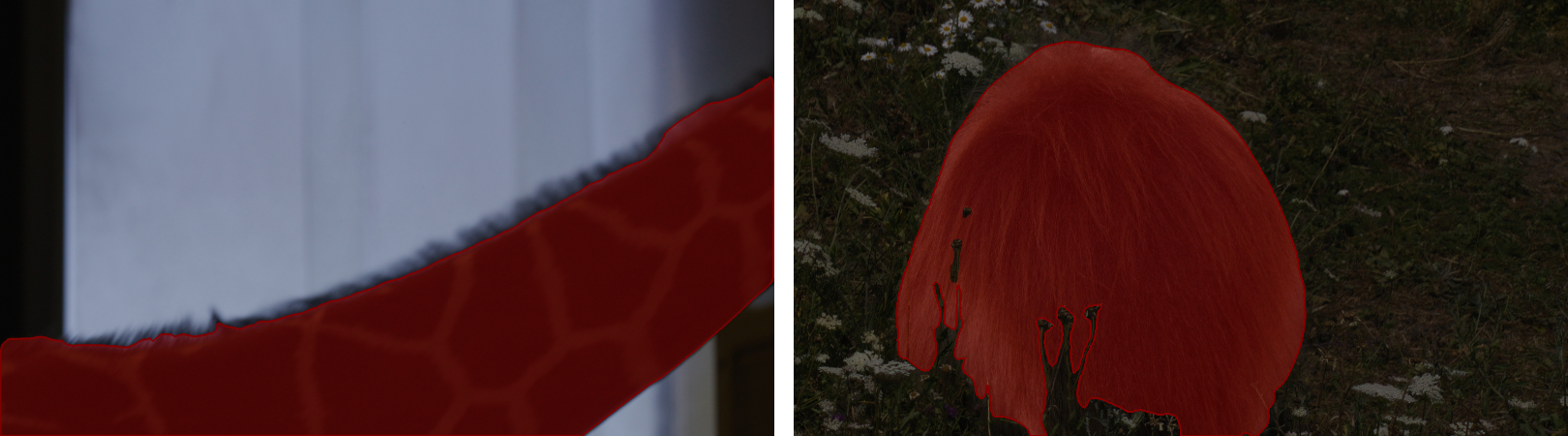}
    \caption{Examples of difficult segmentation scenarios in OLiVES.}
    \label{fig:results_vos_hard}
\end{figure}

To determine the cause of the high VOS performance, we group the objects by size and calculate the common VOS metrics: $\mathcal{J}\&\mathcal{F}$, $\mathcal{J}$ and $\mathcal{F}$. We collect these results using the pre-trained Cutie~\cite{Cheng2024cutie} (to avoid bias to the dataset) and present them in~\cref{tab:suppmat_olives_size_vos_scores}. We find that performances vary for small and medium objects but large objects have significantly higher scores. We can determine that the variance in performance for small and medium objects is due to the low count of both sizes.

However, closer inspection reveals that even when trained on our dataset, Cutie fails to capture fine-grained textures (see~\cref{fig:results_vos_hard}). In these examples, we can see that Cutie struggled to segment the details of the fur of the animals, as well as handling occlusions. While such performance may be sufficient for surveillance applications, it falls short of the precision required for post-production tasks, where high-quality, finely detailed masks are essential.

We conclude that due to the unbalanced size distribution of our dataset (refer to~\cref{fig:dataset_stats} (d)) and the bias of models on large objects, this results in numerically high VOS performances.


\end{document}